\documentclass[10pt,twocolumn,letterpaper]{article}

\usepackage[pagenumbers]{cvpr}  

\usepackage[dvipsnames]{xcolor}

\newcommand{\TODO}[1]{\textbf{\color{red}[TODO: #1]}}
\renewcommand{\TODO}[1]{}

\usepackage{graphicx}
\usepackage{booktabs}
\usepackage{colortbl}
\usepackage{caption}
\usepackage{pifont}
\newcommand{\cmark}{\ding{51}}  
\newcommand{\xmark}{\ding{55}}  

\definecolor{cvprblue}{rgb}{0.21,0.49,0.74}
\usepackage[pagebackref,breaklinks,colorlinks,citecolor=cvprblue]{hyperref}

\title{\textsc{ReJEPA}: A Novel \textsc{J}oint-\textsc{E}mbedding \textsc{P}redictive \textsc{A}rchitecture for Efficient Remote Sensing Image Retrieval}
\author{Shabnam Choudhury \quad Yash Salunkhe \textsuperscript{*} \quad Sarthak Mehrotra \textsuperscript{*}  \quad Biplab Banerjee\\
Indian Institute of Technology Bombay\\
{\tt\small \{choudhury.shabnam6, yashsalunkhe619\}@gmail.com,} \\
{\tt\small \{sarthak2002.mehrotra, getbiplab\}@gmail.com}}

\makeatletter
\renewenvironment{abstract}{
    \centerline{\large\bf Abstract}%
    \vspace{0.5ex}%
    \begin{quote}
    \small
}{
    \end{quote}
    \vspace{-1.5ex}  
}
\makeatother

\begin{document}
\maketitle

\begingroup
\renewcommand\thefootnote{}\footnotetext{\textsuperscript{*}Equal contribution}
\endgroup
\begin{abstract}
The rapid expansion of remote sensing image archives demands the development of strong and efficient techniques for content-based image retrieval (RS-CBIR). This paper presents \textsc{ReJEPA} (Retrieval with Joint-Embedding Predictive Architecture), an innovative self-supervised framework designed for unimodal RS-CBIR. \textsc{ReJEPA} utilises spatially distributed context token encoding to forecast abstract representations of target tokens, effectively capturing high-level semantic features and eliminating unnecessary pixel-level details. In contrast to generative methods that focus on pixel reconstruction or contrastive techniques that depend on negative pairs, \textsc{ReJEPA} functions within feature space, achieving a reduction in computational complexity of 40–60\% when compared to pixel-reconstruction baselines like Masked Autoencoders (MAE).
To guarantee strong and varied representations, \textsc{ReJEPA} incorporates Variance-Invariance-Covariance Regularisation (VICReg), which prevents encoder collapse by promoting feature diversity and reducing redundancy. The method demonstrates an estimated enhancement in retrieval accuracy of 5.1\% on BEN-14K (S1), 7.4\% on BEN-14K (S2), 6.0\% on FMoW-RGB, and 10.1\% on FMoW-Sentinel compared to prominent SSL techniques, including CSMAE-SESD, Mask-VLM, SatMAE, ScaleMAE, and SatMAE++, on extensive RS benchmarks BEN-14K (multispectral and SAR data), FMoW-RGB, and FMoW-Sentinel. Through effective generalization across sensor modalities, \textsc{ReJEPA} establishes itself as a sensor-agnostic benchmark for efficient, scalable, and precise RS-CBIR, addressing challenges like varying resolutions, high object density, and complex backgrounds with computational efficiency. 
\end{abstract}
\vspace{-1mm}
\section{Introduction}
\vspace{-1mm}
Content-based image retrieval (CBIR) plays a crucial role in remote sensing (RS), facilitating efficient retrieval of semantically relevant images from large and continuously expanding archives. The success of CBIR hinges on learning robust image representations that capture high-level semantics. Deep learning (DL)-based methods have become the cornerstone of CBIR, with self-supervised learning (SSL) emerging as a promising approach due to its ability to learn meaningful representations without the need for costly manual annotations \cite{sumbul2021deep, zhao2024dense}. Among SSL methods, contrastive learning \cite{sumbul2022novel, chen2020simple} has shown exceptional promise by capturing semantic features from unannotated datasets. However, its reliance on negative pairs introduces challenges in RS, where different images often belong to the same class, potentially limiting its effectiveness.\par
\begin{figure}
    \centering
    \includegraphics[height = 6.2cm,width = 8.5cm]{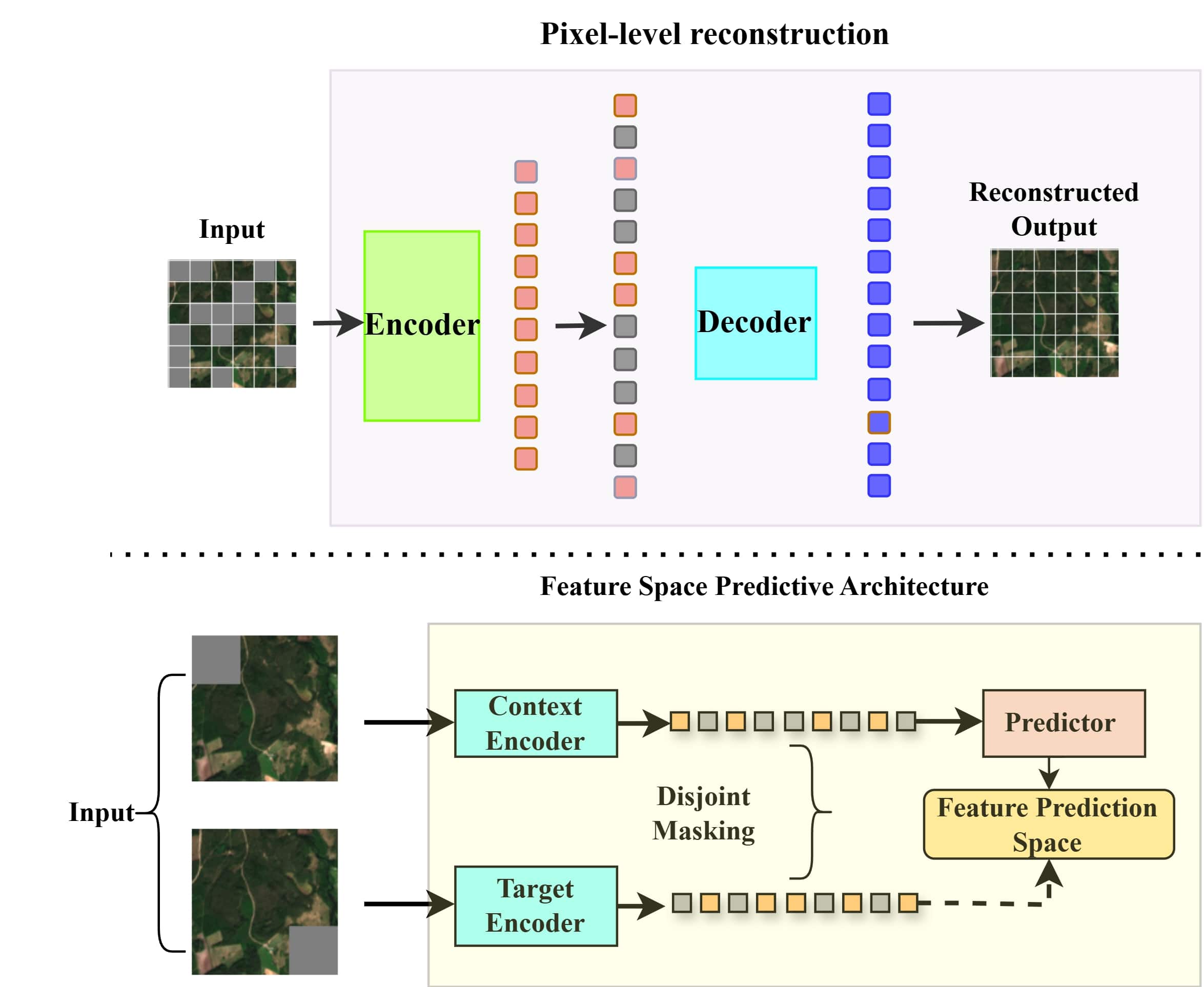}
    \caption{Conceptual illustration of \textsc{ReJEPA} for RS-CBIR. The upper section represents traditional pixel-level reconstruction. The lower section highlights \textsc{ReJEPA} feature-space prediction, The predicted features are then used for retrieval via k-NN}
    \label{Figure1}
    \vspace{-15pt}
\end{figure}
\begin{figure*}[t]
    \centering
    \includegraphics[width=\textwidth, height=6.5cm]{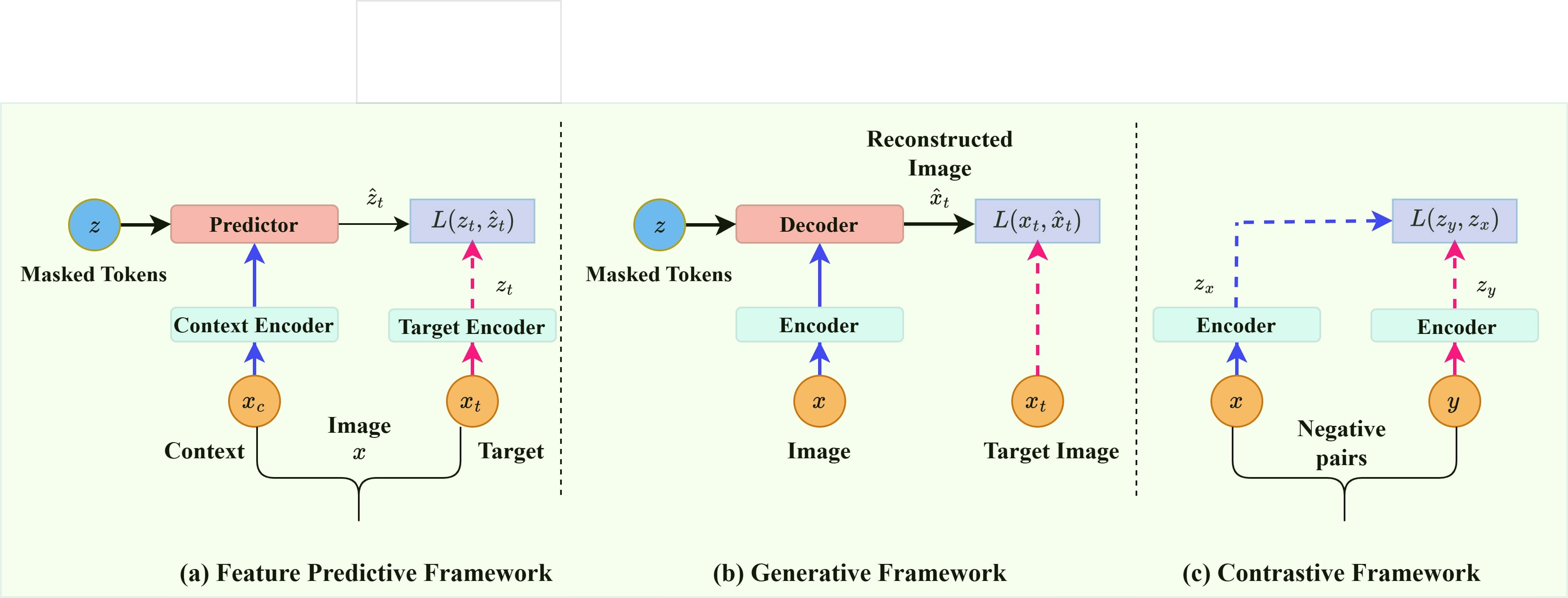}
    \caption{Illustration of self-supervised learning frameworks \cite{li2024predicting} based on their learning objectives. (a) Feature Predictive Framework (e.g., \textsc{ReJEPA}) learns by predicting the feature representation \( \hat{z}_t \) of a target image \( x_t \) from the context representation \( z_c \), using a predictor conditioned on a transformation variable \( z \). (b) Generative Framework reconstructs the pixel-level output \( \hat{x}_t \) from a masked input \( x_t \), employing a decoder for full-image reconstruction. (c) Contrastive Framework learns to align embeddings \( z_x \) and \( z_y \) of semantically similar images while pushing apart representations of negative samples, relying on augmentation-based training.}
    \label{Figure2}
    \vspace{-15pt}
\end{figure*}
Generative SSL methods, such as masked image modeling (MIM) \cite{xie2022simmim, he2022masked}, reconstruct missing pixels to learn feature representations. In RS, MIM-based models like masked autoencoders (MAEs) \cite{he2022masked, cong2022satmae} leverage vision transformers (ViTs) \cite{dosovitskiy2020image} for feature extraction. However, these models emphasize pixel reconstruction, limiting their ability to capture high-level semantics \cite{assran2022masked, yi2022masked}, especially in RS, where diverse resolutions, dense objects, and environmental noise pose additional challenges. To overcome these limitations, we propose \textsc{ReJEPA} (Retrieval with Joint Embedding Predictive Architecture), a novel RS-CBIR framework that replaces pixel reconstruction with feature-space prediction. Built on joint-embedding predictive architectures \cite{assran2023self}, \textsc{ReJEPA} predicts high-level feature representations of target regions using spatially distributed context tokens, capturing semantic structures while discarding pixel-level redundancies. This makes it well-suited for RS tasks involving multi-resolution data, complex object distributions, and noisy imaging conditions. \textsc{ReJEPA} has broad applicability across various RS tasks, including land cover and land use mapping, disaster response and damage assessment, precision agriculture, surveillance, and climate change monitoring. For instance, in disaster response, \textsc{ReJEPA} can quickly retrieve relevant pre- and post-disaster images, aiding in damage assessment and recovery efforts. Similarly, precision agriculture can retrieve images with similar crop health or soil conditions, supporting crop monitoring and yield prediction. Its ability to generalize across diverse sensor modalities (RGB, multispectral, SAR) makes it well-suited for cross-sensor data fusion, addressing complex challenges in remote sensing.

To further enhance \textsc{ReJEPA}’s performance and address the risk of encoder collapse, where embeddings become constant or non-informative, we incorporate Variance-Invariance-Covariance Regularization (VICReg) \cite{bardes2021vicreg}. VICReg employs two regularization terms: (1) maintaining the variance of each embedding dimension above a threshold, ensuring diversity in the learned features, and (2) decorrelating variables to minimize redundancy in the embeddings. These enhancements are crucial for capturing the complex and diverse characteristics of RS imagery. By combining joint-embedding predictive design with VICReg, \textsc{ReJEPA} produces semantically rich and robust representations, achieving faster convergence and superior performance in CBIR tasks. In this work, we propose the following key contributions:

\begin{itemize}
 \item We introduce \textsc{ReJEPA}, the first joint-embedding predictive framework for RS-CBIR, which learns high-level semantic representations without pixel-level reconstruction or reliance on negative pairs, reducing computational complexity by 40–60\%.
\item We incorporate Variance-Invariance-Covariance Regularization (VICReg) to prevent encoder collapse, ensuring robust, diverse, and non-redundant representations tailored for RS-CBIR.
\item We demonstrate that \textsc{ReJEPA} generalizes effectively across different sensor modalities (RGB, multispectral, and SAR), establishing itself as a sensor-agnostic benchmark for efficient and scalable RS-CBIR.
\end{itemize}
\vspace{-3mm}
\section{Related Work}
\vspace{-1mm}
\subsection{Self-Supervised Learning for Remote Sensing}
\vspace{-1mm}
Self-supervised learning (SSL) has revolutionized remote sensing (RS) by alleviating the dependency on large-scale annotated datasets \cite{wang2022self}. Existing SSL methods in RS predominantly fall into contrastive and generative paradigms, as illustrated in Figure \ref{Figure2}. The research landscape of SSL-based foundation models for RS closely mirrors this dichotomy \cite{ayush2021geography, bourcier2025learning, tseng2022croco}.

Contrastive methods, depicted in Figure \ref{Figure2}(c) often referred to as Siamese structures, augmentation-based techniques, or joint-embedding predictive architectures (JEPAs), have demonstrated significant promise in RS applications. For instance, studies such as \cite{jung2021contrastive, zhao2020self} utilized spatial neighbors as augmented data to create positive pairs, while \cite{li2025masked} incorporated random rotations (90°, 180°, and 270°) to enhance data variability. Other efforts, such as \cite{li2021geographical}, exploited geographical vegetation as augmentation cues. Additionally, some approaches predict missing modalities from available data to improve multi-modal representation learning \cite{astruc2025omnisat, fuller2024croma}. Despite these successes, contrastive methods face challenges in RS due to their reliance on negative pairs, which may inadvertently include semantically similar samples, thereby limiting their effectiveness. 

Generative SSL methods,  shown in Figure \ref{Figure2}(b) particularly masked image modeling (MIM), have gained widespread recognition for their ability to learn semantic representations by reconstructing masked regions of an image. MIM-based models, such as masked autoencoders (MAEs) \cite{he2022masked}, have been extensively adapted for RS tasks. These adaptations incorporate domain-specific properties, such as scale invariance \cite{reed2023scale}, temporal information \cite{cong2022satmae}, and temporal invariance \cite{manas2021seasonal}. Advanced generative models, including RingMo \cite{sun2022ringmo}, billion-scale MAE \cite{cha2023billion}, and VITAE \cite{wang2022advancing}, have pushed the boundaries of MIM by scaling the framework to larger datasets and architectures. Generative models in RS leverage unique properties of Earth observation data through spectral \cite{cong2022satmae}, temporal \cite{dumeur2024self}, and spatiotemporal \cite{moor2023foundation} masking strategies.
\vspace{-1mm}
\subsection{Feature Predictive Architectures}
\vspace{-1mm}
Figure \ref{Figure2}(a) illustrates predictive architectures as a promising paradigm in self-supervised learning (SSL), shifting from pixel-level reconstruction to high-level semantic feature prediction. Unlike contrastive or generative methods, which rely heavily on negative pairs or pixel-level reconstruction, feature predictive approaches aim to capture meaningful semantic relationships by leveraging contextual information from the data itself. At the core of this paradigm lies the joint-embedding predictive architecture (JEPA) \cite{lecun2022path}, which employs a Siamese encoder-predictor design to infer missing information in feature space.

JEPAs have been successfully implemented across various modalities, including audio \cite{baevski2022data2vec}, image \cite{zhou2021ibot, oquab2023dinov2, assran2023self}, and text \cite{baevski2023efficient}. These approaches have demonstrated that predicting in representation space leads to versatile representations that perform well in downstream tasks such as linear probing and low-shot adaptation \cite{assran2022masked, assran2023self, oquab2023dinov2}, while offering significant efficiency gains during pretraining compared to pixel-level reconstruction \cite{assran2023self, baevski2022data2vec}. Additionally, feature-predictive methods exhibit competitive performance in end-to-end fine-tuning for image, audio, and text domains \cite{baevski2023efficient, baevski2022data2vec}. Feature-predictive models can also incorporate contrastive objectives for improved stability and representation quality \cite{baevski2022data2vec}.

By bypassing the need for complex data augmentations or decoder networks, JEPAs are particularly well-suited for multi-modal applications, such as Earth observation, where data diversity and scale present unique challenges. For example, SAR-JEPA \cite{li2024predicting} introduced the application of JEPA concepts, focusing specifically on SAR data. JEPA architectures handle diverse sensor modalities (e.g., RGB, multispectral, SAR) without requiring modality-specific designs, adapt to varying resolutions and scales, and are resilient to noise and complex backgrounds. By avoiding pixel-level reconstruction, JEPA ensures computational efficiency and scalability, making it ideal for tasks like RS-CBIR, where robust and versatile representations are critical.

We propose \textsc{ReJEPA}, a novel JEPA-based framework designed specifically for RS-CBIR, which predicts in feature space. \textsc{ReJEPA} learns robust, high-level semantic representations while addressing the unique challenges of RS imagery, including varying sensor types, resolutions, and spatial complexities.

\begin{figure*}[t]
    \centering
    \includegraphics[width=0.9\textwidth]{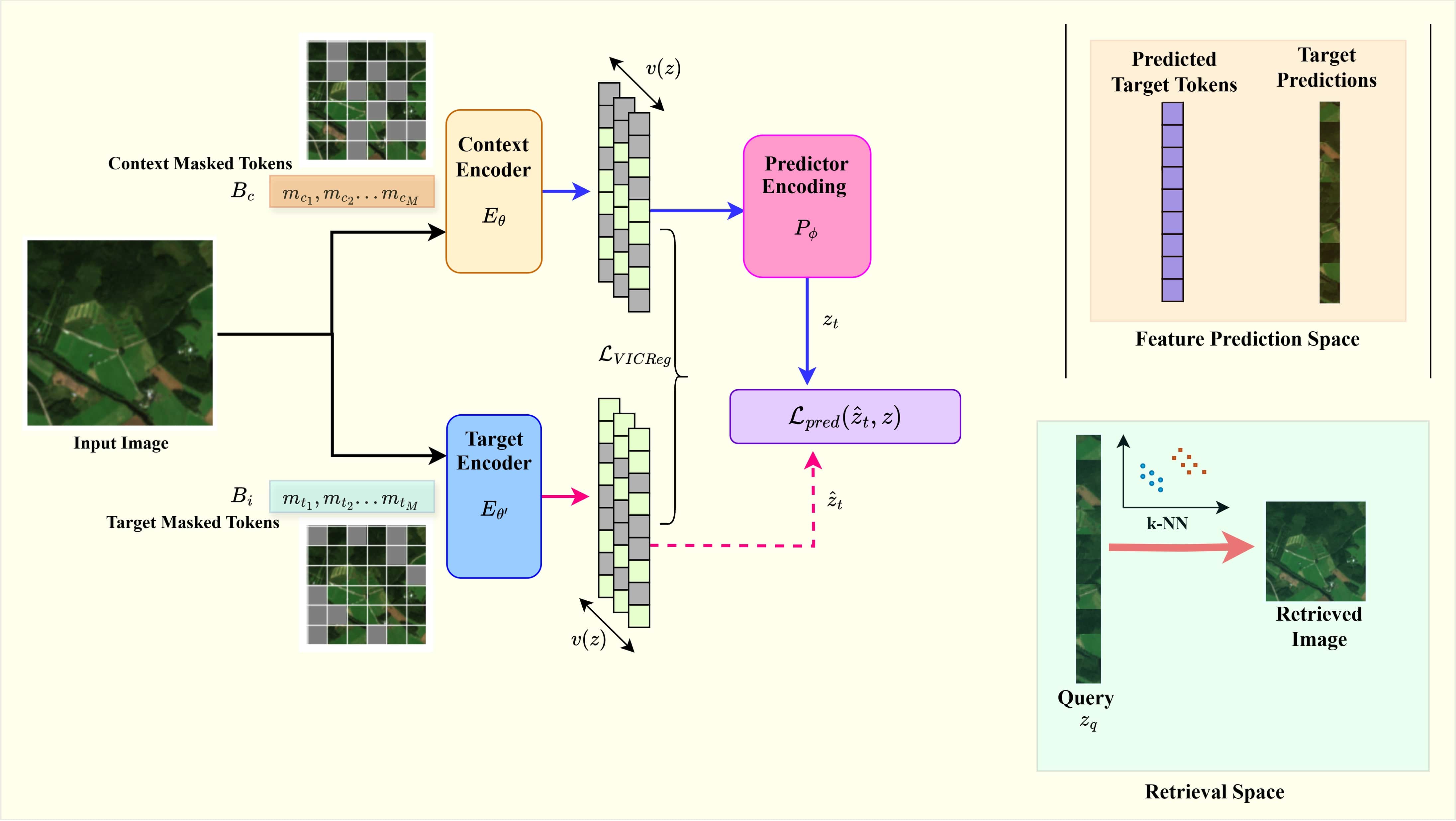}
    \caption{Architecture of \textsc{ReJEPA}, employing a disjoint masking strategy, where context masked tokens (\( B_c \)) and \textbf{target masked tokens} (\( B_i \)) are independently sampled to ensure non-overlapping feature learning. The context encoder (\(E_{\theta}\)) processes unmasked regions to extract spatial representations, while the target encoder (\(E_{\theta'}\)) encodes the masked target patches, generating reference embeddings. The predictor (\(P_{\phi}\)) estimates the missing target representations (\(\hat{z}_t\)) from the context encoding, conditioned on positional tokens. VICReg regularization enforces feature variance (\(v(z)\)), ensuring diverse and stable representations. The retrieval pipeline employs \textbf{k-NN search} on the learned feature space, retrieving semantically similar images from the archive.}
    \label{Figure3}
    \vspace{-10pt}
\end{figure*}
\vspace{-3mm}
\section{Method : \textsc{ReJEPA}}
\vspace{-1mm}
\subsection{Problem Definition}
\vspace{-1mm}
Let \(\mathcal{X} = \{ x_i \}_{i=1}^{N} \) be a remote sensing (RS) image archive of \( N \) images. The goal of remote sensing content-based image retrieval (RS-CBIR) is to retrieve images from \(\mathcal{X}\) that are semantically similar to a query image \( x_q \in \mathcal{X} \). To achieve this, we propose \textsc{ReJEPA}, a joint-embedding predictive architecture (JEPA) based framework \cite{assran2023self, lecun2022path}, as shown in Figure \ref{Figure1}. \textsc{ReJEPA} learns representations by predicting the feature representation of a target image \( x_t \) from a context image \( x_c \), conditioned on a transformation variable \( z \):
\vspace{-1mm}
\begin{equation}
    \hat{z}_t = P_{\phi}(z_c, z_t),
\end{equation}
\vspace{-1mm}
where 
\vspace{-1mm}
\begin{equation}
    z_c = E_{\theta}(x_c), \quad z_t = E_{\theta'}(x_t),
\end{equation}
and \( z \) encodes the spatial and spectral relationships relationship between \( x_c \) and \( x_t \). For retrieval, the representation of a query image \( x_q \) is computed as:
\vspace{-1mm}
\begin{equation}
    z_q = E_{\theta}(x_q),
\end{equation}

and the nearest neighbors \( z_q \) are retrieved in feature space.
\vspace{-1mm}

\subsection{\textbf{\textsc{ReJEPA}} Architecture and Predictive Learning}
\vspace{-1mm}
We introduce \textsc{ReJEPA}, illustrated in Figure \ref{Figure3}, a novel framework for remote sensing content-based image retrieval (RS-CBIR) built on the principles of the feature predictive architectures. The core idea of \textsc{ReJEPA} is to learn semantically meaningful representations by predicting the feature representation of a target image from a context image, conditioned on a transformation variable $z$. This approach shifts the focus from pixel-level reconstruction (as in generative models) or contrastive learning (which relies on negative pairs) to feature-space prediction, enabling the model to capture high-level semantic information while discarding irrelevant pixel-level details. The overall architecture consists of three key components:

\begin{itemize}
    \item \textbf{Context and Target Encoders} (\( E_{\theta}, E_{\theta'} \)): Two encoders process the context and target images separately to generate meaningful feature representations. The \textit{context encoder} extracts spatially distributed features while disregarding masked regions, whereas the \textit{target encoder} reconstructs the missing regions by encoding target patches into an abstract feature space.
    
    \item \textbf{Predictor Network with Learned Transformations} (\( P_{\phi} \)): Learns to predict the representation of masked target embeddings from the context embeddings while encoding transformation cues. By leveraging the context representation and an implicit transformation variable, the predictor estimates the feature distribution of the unseen target, allowing \textsc{ReJEPA} to model high-level semantics beyond pixel-wise reconstruction.
    
    \item \textbf{Feature Regularization via VICReg}: Ensures robust and diverse representations by enforcing variance preservation, decorrelation, and invariance constraints, mitigating encoder collapse while improving feature discrimination in RS-CBIR.
\end{itemize}
\vspace{-1.5mm}
\subsection{Context and Target Encoders}
\vspace{-1mm}
\noindent\textbf{Context Encoder (\( E_{\theta} \))}: A disjoint random masking strategy is employed, where the context mask (\( B_c \)) and target mask (\( B_i \)) are sampled independently, ensuring that no overlapping regions exist between the context and target tokens. This prevents information leakage and enforces a stronger predictive learning objective. The context tokens, masked by \( B_c \), retain only spatially distributed visible regions and are then processed through \( E_{\theta} \) to generate their feature representation:
\vspace{-1mm}
\begin{equation}
    z_c = E_{\theta}(x_c) = \{z_{c1}, z_{c2}, ..., z_{cM}\}
\end{equation}
where \( z_c \) captures non-overlapping contextual features necessary for downstream prediction.
\vspace{2mm}\\
\noindent\textbf{Target Encoder (\( E_{\theta'} \))}: The target encoder \( E_{\theta'} \) processes the masked regions (target patches) independently from the context encoder to generate feature representations that the model aims to predict. A disjoint target mask \( B_i \) is applied, ensuring that target tokens do not overlap with the context tokens, enforcing a stricter predictive setup. The target-encoded representation is then obtained as follows:
\vspace{-1mm}
\begin{equation}
    z_t = E_{\theta'}(x_t) = \{z_{t1}, z_{t2}, ..., z_{tN}\}
\end{equation}
\vspace{-2mm}
\subsection{Predictor Network }
\vspace{-1mm}
The prediction module (\( P_{\phi} \)) in \textsc{ReJEPA} is responsible for estimating the feature representations of masked target regions using only the available contextual features. Given the output of the context encoder \( z_c \), \( P_{\phi} \) learns to infer the representations of \( M \) masked target tokens, ensuring that the model captures high-level semantic relationships rather than low-level pixel dependencies. For a given target tokens representation \( z_t^{(i)} \), associated with a target mask \( B_i \), the predictor network \( P_{\phi}(\cdot, \cdot) \) takes as input:
\begin{itemize}
    \item The context representation \( z_c \), extracted from unmasked image regions.
    \item A set of mask tokens \( \{ m_j \}_{j \in B_i} \) corresponding to the masked target patches.
\end{itemize}
The predictor then estimates the target encoder token representation as:
\begin{equation}
    \hat{z}_t^{(i)} = P_{\phi}(z_c, \{ m_j \}_{j \in B_i})
\end{equation}
where \( \hat{z}_t^{(i)} \) represents the predicted embeddings for the \( i \)-th target token.
\vspace{2mm}\\
\noindent\textbf{Mask Token Parameterization}: The prediction process relies on a set of learnable mask tokens. The mask tokens are parameterized by a shared learnable vector, ensuring that the model generalizes well across different masking patterns. Each mask token is enriched with a positional embedding, allowing the predictor to retain spatial information and ensure sensor-invariant feature learning.
Since we predict \( M \) target tokens, we apply the predictor network \( M \) times, each time conditioning on the corresponding mask tokens for the target locations. This results in a set of predictions:
\begin{equation}
    \{ \hat{z}_t^{(1)}, \hat{z}_t^{(2)}, \dots, \hat{z}_t^{(M)} \}
\end{equation}
which are compared with the actual target representations \( \{ z_t^{(1)}, z_t^{(2)}, \dots, z_t^{(M)} \} \) during training.
\vspace{-1mm}
\subsection{Feature Regularization via VICReg}
\vspace{-1mm}
A major challenge in self-supervised learning is representation collapse, where the encoder outputs degenerate feature representations, mapping all inputs to a constant or non-informative vector. This issue is particularly critical in joint-embedding architectures, such as \textsc{ReJEPA}, where contrastive negative pairs are absent. To mitigate this, we integrate Variance-Invariance-Covariance Regularization (VICReg) \cite{bardes2021vicreg}, which enforces stability and diversity in learned feature representations, \( z \in \mathbb{R}^{d} \), where \( d \) represents the dimensionality of the feature embedding space. Given a batch of \( n \) feature representations, VICReg is formulated by applying three key constraints.
\begin{itemize}
    \item \textbf{Variance Regularization:} Ensures feature diversity by preventing collapse into a single point, maintaining a minimum variance threshold across feature dimensions:
    \begin{equation}
        v(z) = \frac{1}{d} \sum_{j=1}^{d} \max(0, \gamma - \sqrt{\text{Var}(z_j) + \epsilon}),
    \end{equation}
    where \( \gamma \) controls the variance threshold, and \( \epsilon \) stabilizes training.

    \item \textbf{Covariance Regularization:} Reduces redundancy by decorrelating feature dimensions, ensuring each captures distinct semantic information:
    \begin{equation}
        c(z) = \frac{1}{d} \sum_{i \neq j} \left[ \frac{1}{n - 1} \sum_{i=1}^{n} (z_i - \bar{z}) (z_i - \bar{z})^T \right]^2_{i,j},
    \end{equation}
    where \( \bar{z} \) represents the mean embedding.

    \item \textbf{Invariance Regularization:} Encourages consistency between embeddings from different transformations of the same image, ensuring robustness:
    \begin{equation}
        \mathcal{L}_{\text{inv}} = \frac{1}{n} \sum_{i=1}^{n} ||z_i - z_i' ||^2,
    \end{equation}
    where \( z_i \) and \( z_i' \) are embeddings from different augmented views of the same image.
\end{itemize}
\vspace{2mm}
\noindent\textbf{VICReg Integration in \textsc{ReJEPA}}: VICReg regularizes both context and target representations, preventing trivial solutions while enhancing retrieval distinctiveness through feature decorrelation. By enforcing statistical constraints, it ensures diverse, independent embeddings, enabling generalization across sensor modalities (RGB, SAR, multispectral). This regularization stabilizes training, prevents degenerate representations, and improves convergence and retrieval efficiency in large-scale RS-CBIR.
\begin{table*}
    \centering
    \renewcommand{\arraystretch}{1}
    \setlength{\tabcolsep}{8pt}
    \caption{F1-score (\%) comparison of self-supervised models for RS-CBIR on BEN-14K, FMoW-RGB, and FMoW-Sentinel datasets. \textsc{ReJEPA} achieves the best retrieval performance with reduced computational complexity.}
    \begin{tabular}{l|c|cc|c|c}
        \hline
        \rowcolor[HTML]{5D6D7E}  
        \textcolor{white}{\textbf{Model Name}} & \textcolor{white}{\textbf{\#Params (M)}} & \multicolumn{2}{c|}{\textcolor{white}{\textbf{BEN-14K}}} & \textcolor{white}{\textbf{FMoW-RGB}} & \textcolor{white}{\textbf{FMoW-Sentinel}} \\
        \rowcolor[HTML]{5D6D7E}  
        & & \textcolor{white}{\textbf{S1$\rightarrow$S1}} & \textcolor{white}{\textbf{S2$\rightarrow$S2}} & & \\
        \hline
        \rowcolor[HTML]{EAF2F8} MAE \cite{he2022masked} & 224.87 & 60.81 & 72.04 & 58.73 & 61.77 \\
        MAE-RVSA \cite{wang2022advancing} & 227.75 & 55.40 & 71.47 & 55.26 & 60.28 \\
        \rowcolor[HTML]{EAF2F8} SatMAE \cite{cong2022satmae} & 329.40 & 70.86 & \textbf{78.71} & 61.85 & 56.63 \\
        SatMAE++ \cite{noman2024rethinking} & 329.14 & 67.29 & \textbf{76.48} & 60.09 & 57.75 \\
        \rowcolor[HTML]{EAF2F8} SS-CMIR \cite{sumbul2022novel} & 259.07 & 68.07 & 70.54 & 66.71 & 63.46 \\
        Scale-MAE \cite{reed2023scale} & 284.35 & 62.73 & NA & 64.26 & 69.56 \\
        Mask-VLM \cite{kwon2022masked} & 225.82 & 68.10 & 71.02 & 61.52 & 65.23 \\
        \rowcolor[HTML]{EAF2F8} CSMAE-SESD (Disjoint) \cite{hackstein2024exploring} & 210.64 & 70.62 & 39.01 & 68.42 & 57.13 \\
        \hline
        \rowcolor[HTML]{A3E4D7}  
        \textbf{\textsc{ReJEPA}} & \textbf{197.09} & \textbf{76.38} & 75.42 & \textbf{73.53} & \textbf{75.87} \\
        \hline
    \end{tabular}
    \label{tab1}
\end{table*}

\vspace{-1mm}
\subsection{Training Objective}
\vspace{-1mm}
The training objective of \textsc{ReJEPA} is designed to ensure both accurate predictive learning and stable feature representations by combining a prediction loss $(\mathcal{L}_{\text{pred}})$ that aligns context and target representations with a regularization loss $(\mathcal{L}_{\text{VICReg}})$ that prevents feature collapse and redundancy. The prediction loss minimizes the discrepancy between the predicted target representation \( \hat{z}_t^{(i)} \) and the actual target representation \( z_t^{(i)} \), using an $L_2$ loss formulation:
\vspace{-1mm}
\begin{equation}
    \mathcal{L}_{\text{pred}} = \frac{1}{M} \sum_{i=1}^{M} || \hat{z}_t^{(i)} - z_t^{(i)} ||_2^2,
\end{equation}
where \( M \) is the number of masked target tokens. To maintain feature diversity and avoid degenerate solutions, \textsc{ReJEPA} incorporates a feature regularization loss that consists of three key terms: variance regularization \( v(z) \), covariance regularization \( c(z) \), and an invariance term \( \mathcal{L}_{\text{inv}} \). These regularization terms are combined as follows:
\vspace{-1mm}
\begin{equation}
    \mathcal{L}_{\text{VICReg}} = \lambda_v v(z) + \lambda_c c(z) + \lambda_i \mathcal{L}_{\text{inv}},
\end{equation}
where \( \lambda_v, \lambda_c, \lambda_i \) are hyperparameters controlling the relative influence of each constraint. The final training objective of \textsc{ReJEPA} is a weighted sum of the prediction and regularization losses:
\vspace{-1mm}
\begin{equation}
    \mathcal{L} = \mathcal{L}_{\text{pred}} + \mathcal{L}_{\text{VICReg}},
\end{equation}
where \( \alpha \) determines the balance between predictive accuracy and feature regularization. Once trained, the model generates latent feature representations for all images in the archive. Given a query image \( x_q \), its corresponding representation is obtained via the context encoder, \( E_{\theta}(x_q) \). To retrieve the most relevant images, a \textit{k}-nearest neighbors (\textit{k}-NN) search is performed in the learned feature space.

\begin{figure*}
    \centering
    \includegraphics[width=\textwidth]{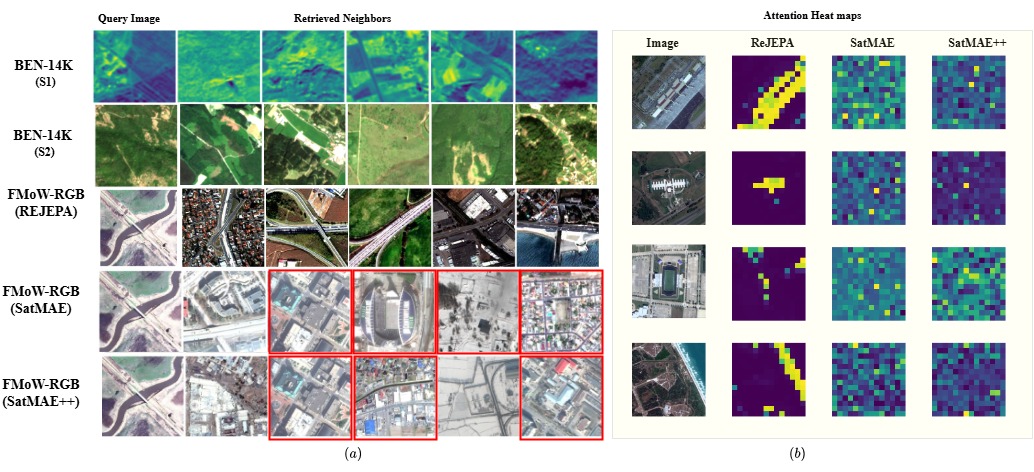}
    \caption{Qualitative retrieval and attention visualization results of \textsc{ReJEPA}. (a) Retrieval results on BEN-14K and FMoW datasets. On FMoW, \textsc{ReJEPA} retrieves more semantically coherent images than SatMAE and SatMAE++, particularly in complex scenes with high intra-class variance. (b) Context encoder attention heatmaps comparing \textsc{ReJEPA} with SatMAE and SatMAE++. \textsc{ReJEPA} attends to key structural regions while SatMAE and SatMAE++ exhibit more dispersed and less discriminative attention patterns, highlighting the advantage of feature-space prediction for remote sensing retrieval. \textcolor{red}{Red} colored boxes show wrong retrieval.}
    \label{Figure4}
    \vspace{-10pt}
\end{figure*}
\vspace{-1mm}
\section{Experiments}
\vspace{-1mm}
\subsection{Datasets}
\vspace{-1mm}
To evaluate the performance of \textsc{ReJEPA}, we conduct experiments on three large-scale publicly available remote sensing datasets spanning multiple sensor modalities, including SAR, multispectral, and RGB satellite imagery.
\vspace{2mm}\\
\noindent\textbf{BigEarthNet-14K (BEN-14K)}  
This dataset comprises 14,832 Sentinel-1 SAR and Sentinel-2 multispectral image pairs, annotated with 19 land cover categories. The SAR images include VV and VH polarization bands, while multispectral bands at 10m and 20m resolution are standardized via bicubic interpolation.
\vspace{2mm}\\
\noindent\textbf{fMoW-RGB}: The fMoW-RGB dataset \cite{christie2018functional} consists of high-resolution satellite images categorized into 62 different classes, designed primarily for classification tasks. It contains approximately 363,000 training images and 53,000 test images.
\vspace{2mm}\\
\noindent\textbf{fMoW-Sentinel}: Derived from fMoW-RGB, fMoW-Sentinel \cite{cong2022satmae} extends the dataset by incorporating Sentinel-2 multispectral imagery. It maintains the same 62-class taxonomy while offering a significantly larger collection of images, comprising 712,874 training samples, 84,939 validation samples, and 84,966 test images. 
\vspace{-1mm}
\subsection{Implementation and Evaluation Details}
\vspace{-1mm}
\textsc{ReJEPA} employs ViT-B/16 \cite{dosovitskiy2020image} architectures for the context encoder, target encoder, and predictor. The context and target encoders follow the standard ViT design, with the context encoder processing unmasked regions and the target encoder focusing on masked regions to generate target representations. The predictor is a lightweight ViT with 384-dimensional embedding and a depth of 12 layers, matching the number of attention heads in the context encoder. During evaluation, the target encoder’s output is average-pooled to produce a global image representation for retrieval tasks. We optimize \textsc{ReJEPA} using AdamW \cite{loshchilov2017decoupled} with an initial learning rate of \(10^{-4}\), linearly increased to \(10^{-3}\) over the first 15 epochs and decayed to \(10^{-6}\) using a cosine schedule. The weight decay is gradually ramped up from 0.04 to 0.4 throughout training. The target encoder weights are updated via an exponential moving average (EMA) \cite{assran2022masked, caron2021emerging}, starting with a momentum of 0.996 and linearly increasing to 1 throughout training. \textsc{ReJEPA} employs a random disjoint masking strategy with an optimal masking ratio of 0.25, distinguishing itself from conventional predictive learning approaches \cite{assran2023self}. Context and target masks are independently sampled for each image in the mini-batch size of 128, ensuring diverse masking patterns. Additionally, the coefficients of VICReg $[ \lambda_v,  \lambda_c,  \lambda_i]$, employed to prevent degenerate solutions, are set to [25, 25, 1] as followed by \cite{bardes2021vicreg}.

To evaluate the performance of \textsc{ReJEPA} in content-based image retrieval (CBIR) tasks, we adopt the F1 score as the primary metric. The F1 score, defined as the harmonic mean of precision and recall, provides a balanced measure of retrieval accuracy by considering both the relevance of retrieved images (precision) and the completeness of the retrieval process (recall). The retrieval performance is evaluated based on the top-10 retrieved images for each query, where both the query and archive images are encoded into the feature space using the encoder. The encoded archive images serve as a retrieval bank, from which the most relevant images are retrieved based on the query, similar to CSMAE \cite{hackstein2024exploring}.
\vspace{-1mm}
\subsection{Retrieval Performance}
\vspace{-1mm}
We assess the effectiveness of \textsc{ReJEPA} by benchmarking it against state-of-the-art self-supervised learning (SSL) models for content-based image retrieval (RS-CBIR). The competing methods include MAE \cite{he2022masked}, MAE-RVSA \cite{wang2022advancing}, SatMAE \cite{cong2022satmae}, SatMAE++\cite{noman2024rethinking}, SS-CMIR \cite{sumbul2022novel}, Scale-MAE \cite{reed2023scale}, MaskVLM \cite{kwon2022masked}, and CSMAE-SESD \cite{hackstein2024exploring}. To ensure a fair comparison, all models were trained for the same number of epochs. The retrieval performance is evaluated across three diverse datasets—BEN-14K \cite{sumbul2021bigearthnet}, FMoW-RGB\cite{christie2018functional}, and FMoW-Sentinel \cite{cong2022satmae}, covering a range of SAR, multispectral, and high-resolution optical imagery. Table \ref{tab1} shows that \textsc{ReJEPA} surpasses existing SSL-based retrieval models (of same backbone ViT-B/16) while eliminating pixel reconstruction overhead, achieving superior performance with significantly lower computation and better scalability for large-scale RS data. Compared to parameter-heavy MAE-based models such as SatMAE, SatMAE++, Scale-MAE, and MAE, \textsc{ReJEPA} reduces the model size by 10–40\%.
\vspace{-1pt}

On BEN-14K, \textsc{ReJEPA} achieves an overall 4.5\% improvement in F1-score, consistently excelling in S1 $\rightarrow$ S1 and S2 $\rightarrow$ S2 retrieval, surpassing MAE and SS-CMIR by 4–6\% while maintaining efficiency across SAR and multispectral modalities. Notably, \textsc{ReJEPA} performs 22\% better for S1 and 2\% better for S2 than all MAE-based setups, except for SatMAE and SatMAE++ in the S2 setup, where these models benefit from extensive scale-specific optimizations. Additionally, Scale-MAE \cite{reed2023scale} cannot be utilized with multi-spectral data (S2 $\rightarrow$ S2) due to its reliance on Ground Sample Distance Positional Encoding (GSDPE), which constrains its application to RGB channels only.

For a fair comparison, we take CSMAE-SESD (Disjoint) as a baseline model since it follows a similar non-overlapping masking strategy, aligning with our experimental setup. In FMoW-RGB, where scene complexity and intra-class variance pose challenges, \textsc{ReJEPA} improves retrieval by 8.7\% over SS-CMIR and 17.2\% over SatMAE, demonstrating its effectiveness in learning robust semantic representations. Similarly, on FMoW-Sentinel, it generalizes effectively despite spectral distortions and resolution mismatches, achieving a 13.2\% improvement over SS-CMIR and 10.1\% over MaskVLM, highlighting its adaptability to multispectral satellite data.
\vspace{-1mm}
\subsection{Visualization Analysis}
\vspace{-1mm}
To illustrate the effectiveness of \textsc{ReJEPA} in RS-CBIR, we present retrieval results on BEN-14K and FMoW datasets, along with comparative retrieval performance and attention visualizations.  Figure \ref{Figure4}(a) shows top-5 retrievals, demonstrating our model’s ability to retrieve semantically relevant images across varying resolutions and sensor modalities. Unlike generative methods that emphasize pixel similarity, our feature-predictive learning prioritizes semantic consistency, ensuring structurally coherent retrievals. For FMoW datasets, we conduct a direct comparison against SatMAE and SatMAE++, on a particular query like \textit{roadbridge}. As shown in Figure \ref{Figure4}(a), while these models achieve strong performance, they tend to focus on texture-based similarities rather than structural and semantic relationships. On the contrary, \textsc{ReJEPA}, by leveraging feature-space prediction, retrieves semantically meaningful scenes rather than relying on surface-level pixel cues, achieving higher retrieval accuracy even in complex urban and natural environments. Figure \ref{Figure4}(b) visualizes attention heatmaps from the context encoder. Unlike pixel-reconstruction methods, which disperse attention across the entire image, \textsc{ReJEPA} selectively focuses on salient geographic structures, such as roads, vegetation patches, and water bodies, while suppressing background clutter. This spatial selectivity enhances retrieval accuracy, particularly in multi-sensor settings where spectral distortions and resolution mismatches pose challenges.
\vspace{-6mm}

\subsection{Ablation Studies}
\vspace{-1mm}
We conducted a series of ablation studies on BEN-14K to analyze the impact of predictor depth, masking strategy, masking ratio, and VICReg regularization on RS-CBIR performance.
\vspace{1mm}\\
\noindent\textbf{Effect of Predictor Depth}: Table \ref{tab:abl_pred_depth} shows that increasing the predictor depth enhances retrieval performance, with 12 layers yielding a 26.5\% and 27.8\% improvement over a shallow 6-layer predictor for S1$\rightarrow$S1 and S2$\rightarrow$S2, respectively. This confirms that deeper predictors enable more expressive feature mapping to capture high-level semantic relationships between context and target representations. 
\vspace{1mm}\\
\noindent\textbf{Impact of Masking Strategy}: As seen in Table \ref{tab:abl_masking_strategy}, the random masking strategy outperforms the conventional multi-block masking \cite{assran2023self} by 6.9\% (S1$\rightarrow$S1) and 7.9\% (S2$\rightarrow$S2), validating the effectiveness of our disjoint random masking approach in learning non-trivial contextual dependencies. Unlike multi-block masking, where structured regions of an image are masked together, random masking increases feature diversity and reduces spurious correlations between context and target regions for RS images.
\begin{table}
    \centering
    \renewcommand{\arraystretch}{1.3}
    \setlength{\tabcolsep}{8pt}
    \caption{Ablation Study on Predictor Depth for RS-CBIR Performance.}
    \begin{tabular}{c|cc}
        \hline
        \rowcolor[HTML]{5D6D7E}  
        \textcolor{white}{\textbf{Depth}} & \textcolor{white}{\textbf{S1$\rightarrow$S1}} & \textcolor{white}{\textbf{S2$\rightarrow$S2}} \\
        \hline
        \rowcolor[HTML]{EAF2F8} 6  & 61.62 & 59.01 \\
        8  & 64.96 & 62.61 \\
        \rowcolor[HTML]{EAF2F8} 10 & 67.61 & 66.82 \\
        12 & \textbf{76.38} & \textbf{75.42} \\
        \hline
    \end{tabular}
    \label{tab:abl_pred_depth}
\end{table}

\begin{table}
    \centering
    \renewcommand{\arraystretch}{1.3}
    \setlength{\tabcolsep}{8pt}
    \caption{Ablation Study on Masking Strategy for RS-CBIR Performance.}
    \begin{tabular}{l|cc}
        \hline
        \rowcolor[HTML]{5D6D7E}  
        \textcolor{white}{\textbf{Masking Strategy}} & \textcolor{white}{\textbf{S1$\rightarrow$S1}} & \textcolor{white}{\textbf{S2$\rightarrow$S2}} \\
        \hline
        \rowcolor[HTML]{EAF2F8} Multi-block & {71.45} & {69.84} \\
        Random  & \textbf{76.38} & \textbf{75.42} \\
        \hline
    \end{tabular}
    \label{tab:abl_masking_strategy}
\end{table}
\begin{table}
    \centering
    \renewcommand{\arraystretch}{1.3}
    \setlength{\tabcolsep}{8pt}
    \caption{Ablation Study on Masking Ratio for RS-CBIR Performance.}
    \begin{tabular}{c|cc}
        \hline
        \rowcolor[HTML]{5D6D7E}  
        \textcolor{white}{\textbf{Masking Ratio}} & \textcolor{white}{\textbf{S1$\rightarrow$S1}} & \textcolor{white}{\textbf{S2$\rightarrow$S2}} \\
        \hline
        \rowcolor[HTML]{EAF2F8} 0.25  & \textbf{76.38} & \textbf{75.42} \\
        0.40  & 71.56 & 70.54  \\ 
        \rowcolor[HTML]{EAF2F8} 0.85  & 68.47 & 66.46 \\
        \hline
    \end{tabular}
    \label{tab:abl_masking_ratio}
\end{table}
\vspace{1mm}

\begin{table}
    \centering
    \renewcommand{\arraystretch}{1.3}
    \setlength{\tabcolsep}{8pt}
    \caption{Ablation Study on VICReg Regularization for RS-CBIR Performance.}
    \begin{tabular}{c|cc}
        \hline
        \rowcolor[HTML]{5D6D7E}  
        \textcolor{white}{\textbf{VICReg Applied}} & \textcolor{white}{\textbf{S1$\rightarrow$S1}} & \textcolor{white}{\textbf{S2$\rightarrow$S2}} \\
        \hline
        \rowcolor[HTML]{EAF2F8} \cmark & \textbf{76.38} & \textbf{75.42} \\
        \xmark & 64.51 & 62.65 \\
        \hline
    \end{tabular}
    \label{tab:abl_vicreg}
\end{table}

\noindent\textbf{Impact of Masking Ratio}: Table \ref{tab:abl_masking_ratio} suggests that an optimal masking ratio of 0.25 achieves the best performance, improving retrieval by 11.5\% (S1$\rightarrow$S1) and 13.5\% (S2$\rightarrow$S2) compared to a higher 0.85 masking ratio. 
\vspace{1mm}\\
\noindent\textbf{Effect of VICReg Regularisation}: As shown in Table \ref{tab:abl_vicreg}, incorporating VICReg significantly enhances representation learning, improving retrieval by 18.4\% (S1$\rightarrow$S1) and 20.3\% (S2$\rightarrow$S2). Without VICReg, the model suffers from representation collapse, where embeddings become redundant, reducing retrieval effectiveness. VICReg mitigates this by enforcing feature diversity (variance), ensuring distinct feature dimensions (decorrelation), and maintaining consistency between augmented views (invariance). This leads to more discriminative and compact feature embeddings, directly improving retrieval precision.

\vspace{-1mm}
\section{Takeaways}
\vspace{-1mm}
This work introduces \textsc{ReJEPA}, a novel joint-embedding predictive architecture tailored for efficient and scalable RS-CBIR. By shifting from pixel-level reconstruction to feature-space prediction, the proposed framework addresses critical challenges in remote sensing image retrieval, including multi-sensor generalization, varying spatial resolutions, and complex scene structures. \textsc{ReJEPA} is well-suited for real-world efficient remote sensing applications, including disaster response, land use monitoring, environmental surveillance, and defense intelligence.

{
    \small
    \bibliographystyle{ieeenat_fullname}

}

\maketitlesupplementary

\section{Introduction}
\begin{figure*}[h]
    \centering
    \includegraphics[width=\textwidth]{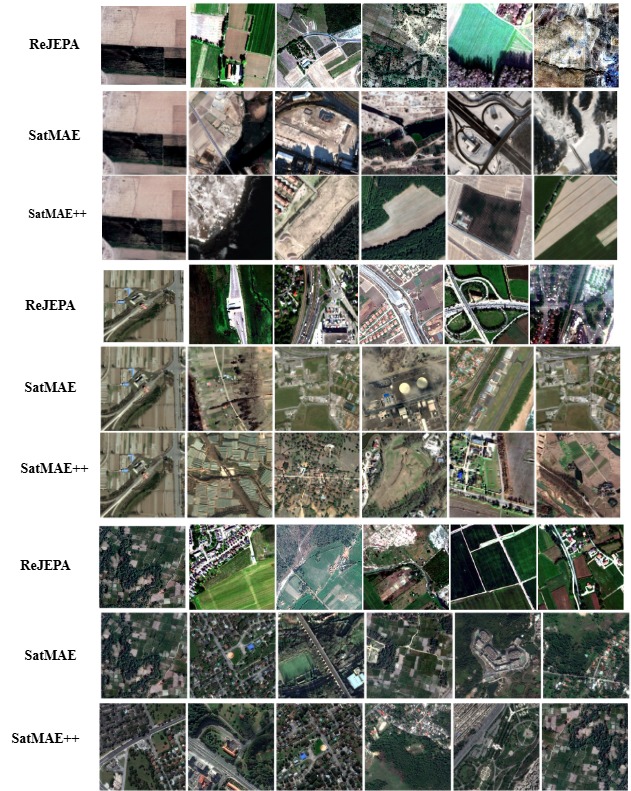}
    \caption{Comparative performance of \textsc{REJEPA} on critical queries with SatMAE and SatMAE++}
    \label{Figure5}
    \vspace{-10pt}
\end{figure*}
We provide additional qualitative analyses to further validate the effectiveness of \textsc{ReJEPA} for remote sensing image retrieval.
\begin{itemize}

    \item In Section \ref{sec: vis}, we present an extensive comparison of \textsc{ReJEPA} against existing methods on more challenging query images, demonstrating its robustness in retrieving semantically relevant results across diverse scenarios. These comparisons reinforce the model’s ability to handle high intra-class variance and complex spatial patterns. 

    \item In Section \ref{sec: attn}, we analyze the attention heatmaps of \textsc{ReJEPA}’s context encoder, showcasing its superior ability to focus on structurally significant regions compared to pixel-reconstruction-based models. This visualization confirms that feature-space prediction not only enhances semantic representation but also achieves better retrieval performance with significantly lower computational overhead. 
\end{itemize}

\begin{figure*}[h]
    \centering
    \includegraphics[width=\textwidth]{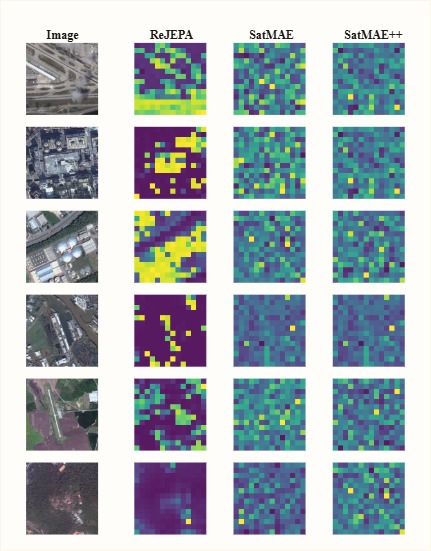}
    \caption{Attention heatmaps of the context encoder \textsc{REJEPA} vs the pixel-wise reconstruction encoders of SatMAE and SatMAE++}
    \label{Figure6}
    \vspace{-10pt}
\end{figure*}

\begin{figure*}[h]
    \centering
    \includegraphics[width=\textwidth]{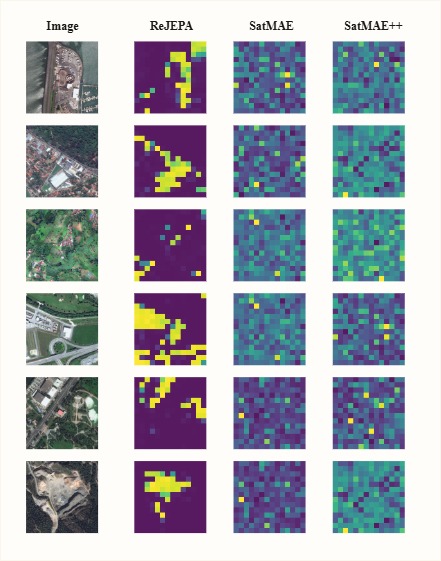}
    \caption{Attention heatmaps of the context encoder \textsc{REJEPA} vs the pixel-wise reconstruction encoders of SatMAE and SatMAE++}
    \label{Figure7}
    \vspace{-10pt}
\end{figure*}

\section{Qualitative Performance}
\label{sec: vis}
To further validate the retrieval effectiveness of \textsc{ReJEPA}, we perform qualitative comparisons against SatMAE and SatMAE++ on challenging query images, focusing on crop fields, a crucial application in remote sensing content-based image retrieval (RS-CBIR). Accurate retrieval of semantically similar agricultural regions is essential for monitoring crop health, assessing land use changes, and analyzing agricultural patterns over time.

 Figure \ref{Figure5} presents retrieval results for crop field queries across the FMoW datasets. \textsc{ReJEPA} consistently retrieves images that exhibit structural and spectral similarities to the query, preserving fine-grained texture and spatial distribution. In contrast, SatMAE\cite{cong2022satmae} and SatMAE++ \cite{noman2024rethinking} often retrieve images with visually dissimilar characteristics due to their reliance on pixel-level reconstruction, which struggles to capture high-level semantic relationships. Notably, \textsc{ReJEPA} achieves greater consistency in identifying homogeneous vegetation areas, correctly retrieving fields with similar crop structures and spectral signatures.

\textbf{Advantages of Feature-Space Prediction:} Unlike pixel-reconstruction-based models, which emphasize low-level pixel alignment, \textsc{ReJEPA} learns feature-space representations that prioritize semantic information. This enables robust retrieval across varying illumination conditions, seasonal changes, and different geographical locations, making it particularly suitable for large-scale agricultural monitor.

\section{Attention Visualisation}
\label{sec: attn}
To further demonstrate the effectiveness of \textsc{ReJEPA}, we present attention heatmaps from the context encoder and compare them with those of SatMAE and SatMAE++. These visualizations provide insights into how feature-space prediction enhances representation learning, leading to more discriminative and semantically rich retrieval performance.

\textbf{Comparison of Attention Patterns:} Figure \ref{Figure6} and \ref{Figure7} shows attention maps for representative remote sensing scenes, including urban areas, agricultural fields, and infrastructure sites. The attention heatmaps reveal distinct differences between \textsc{ReJEPA} and pixel-reconstruction-based models:
\begin{itemize}
    \item \textbf{\textsc{ReJEPA}:} Focuses on high-level semantic structures such as roads, buildings, and vegetation clusters, effectively capturing scene-relevant spatial patterns.
    \item \textbf{SatMAE and SatMAE++:} Display scattered and less interpretable attention distributions, often failing to capture coherent object structures due to their reliance on pixel reconstruction.
\end{itemize}

\end{document}